\documentclass[letterpaper, 10 pt, conference]{ieeeconf}  

\IEEEoverridecommandlockouts                             

\usepackage{graphicx} 
\usepackage{amsmath}
\usepackage{bm}
\usepackage{xcolor}
\usepackage{lipsum}
\usepackage{amssymb}  

\usepackage[style=ieee]{biblatex}

\DeclareSourcemap{
  \maps{
    \map{
      \pertype{article}
      \step[fieldset=url, null]
      \step[fieldset=doi, null]
      \step[fieldset=issn, null]
      \step[fieldset=isbn, null]
      \step[fieldset=note, null]
      \step[fieldset=editor, null]
      \step[fieldset=urldate, null]
      \step[fieldset=file, null]
    }
  }
}
\DeclareSourcemap{
  \maps{
    \map{
      \pertype{inproceedings}
      \step[fieldset=url, null]
      \step[fieldset=doi, null]
      \step[fieldset=issn, null]
      \step[fieldset=isbn, null]
      \step[fieldset=note, null]
      \step[fieldset=editor, null]
      \step[fieldset=urldate, null]
      \step[fieldset=file, null]
    }
  }
}
\DeclareSourcemap{
  \maps{
    \map{
      \pertype{incollection}
      \step[fieldset=url, null]
      \step[fieldset=doi, null]
      \step[fieldset=issn, null]
      \step[fieldset=isbn, null]
      \step[fieldset=note, null]
      \step[fieldset=editor, null]
      \step[fieldset=urldate, null]
      \step[fieldset=file, null]
    }
  }
}

\title{\LARGE \bf
Narrow-Path, Dynamic Walking Using Integrated \\Posture Manipulation and Thrust Vectoring
}

\author{Kaushik Venkatesh Krishnamurthy$^{1}$, Chenghao Wang$^{1}$, Shreyansh Pitroda$^{1}$, Adarsh Salagame$^{1}$, Eric Sihite$^{2}$, \\ Reza Nemovi$^{2}$, Alireza Ramezani$^{1*}$,  Morteza Gharib$^{2}$ 
\thanks{$^{1}$The author is with Department of Electrical and Computer Engineering,
        Northeastern University, Boston, MA, USA  {\tt\small venkateshkrishnamu.k, wang.chengh, pitroda.s, a.salagame@northeastern.edu}}%
\thanks{$^{2}$The author is with the Department of Aerospace Engineering, California Institute of Technology, Pasadena, CA, USA {\tt\small rnemovi, esihite@caltech.edu}%
}
\thanks{$^{*}$Corresponding author {\tt\small a.ramezani@northeastern.edu}}}

\begin{document}

\maketitle
\thispagestyle{empty}
\pagestyle{empty}

\begin{abstract}

This research concentrates on enhancing the navigational capabilities of Northeastern University's Husky, a multi-modal quadrupedal robot, that can integrate posture manipulation and thrust vectoring, to traverse through narrow pathways such as walking over pipes and slacklining. The Husky is outfitted with thrusters designed to stabilize its body during dynamic walking over these narrow paths. The project involves modeling the robot using the HROM (Husky Reduced-Order Model) and developing an optimal control framework. This framework is based on polynomial approximation of the HROM and a collocation approach to derive optimal thruster commands necessary for achieving dynamic walking on narrow paths. The effectiveness of the modeling and control design approach is validated through simulations conducted using Matlab.

\end{abstract}

\section{Introduction}
\label{sec:intro}

\begin{figure}
    \centering
    \includegraphics[width = \linewidth]{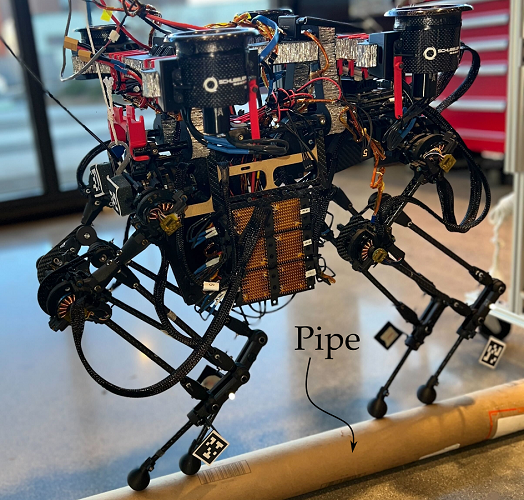}
    \caption{The Northeastern University Husky Carbon walking on a tube.}
    \label{fig:cover-image}
\end{figure}

The quest for versatile robot locomotion has spurred significant advancements in legged locomotion, with numerous bipedal and quadrupedal robots showcasing impressive behaviors across diverse environments \cite{noauthor_robots_nodate,grizzle_progress_nodate,park_finite-state_2013,sreenath_compliant_2011,ramezani_performance_2014}. This study delves into the utilization of a multimodal legged-aerial robot named Northeastern Husky Carbon \cite{ramezani_generative_2021,wang_legged_2023,liang_rough-terrain_2021,sihite_unilateral_2021,sihite_optimization-free_2021, venkatesh_krishnamurthy_towards_2023,salagame_quadrupedal_2023} (illustrated in Fig.~\ref{fig:cover-image}) for narrow path walking. Departing from conventional legged locomotion paradigms, our research explores an innovative concept termed integrated posture manipulation and thrust-vectoring, inspired by avian behaviors \cite{abourachid_hoatzin_2019,gatesy_bipedal_1991}. 

The concept of multimodality and the integration of posture manipulation and thrust-vectoring \cite{sihite_multi-modal_2023} present intriguing unexplored avenues for control design and locomotion strategies. Apart from the obvious scenarios, such as transitioning between legged and aerial locomotion, the fusion of legs and thrusters can lead to other locomotion strategies observed extensively in the animal kingdom. 

For instance, the impressive multimodal capabilities of certain animals in navigating challenging terrains have garnered significant attention. Examples include Chukar birds, which adeptly ascend steep inclines \cite{dial_wing-assisted_2003}, executing agile maneuvers such as rapid walking, leaping, and jumping using both their legs and wings.

Inspiration can also be derived from vertebrate animals that utilize their tails or other inertial appendages to generate momentum and actuate their bodies, although they do not necessarily integrate posture manipulation and fluid-structure interaction. This behavior is observable in various species, such as leopards and other big cats, which rely on their tails for inertial reorientation and balancing on narrow pathways. Additionally, mountain goats utilize their robust neck muscles to secure tight footholds while scaling steep slopes.

Legged-aerial robots, particularly those inheriting dynamic locomotion gaits from animals, grapple with challenges in software and hardware design that make posture manipulation and thrust-vectoring difficult, including limitations in energy density, mass constraints, and under-actuation. In contrast, birds have adeptly addressed such challenges by leveraging lightweight musculoskeletal structures, efficient kinematic trees, and high-energy density muscles, providing the necessary energy density for effective actuation and posture manipulation of their morphing bodies \cite{dial_wing-assisted_2003,tobalske_aerodynamics_2007,peterson_experimental_2011}. These structures and underlying sensory-motor relationships enable animals to efficiently perform posture manipulation and manipulate aerodynamic forces in desired directions, pushing their bodies with dexterity in desired directions—a result of evolutionary refinement that we aim to harness in our designs.

The primary goal of this research is to devise a control design methodology that integrates Husky Carbon's posture and its thrust vectoring capability, facilitated by an array of electric ducted fans affixed to its torso (refer to Section~\ref{sec:husky-overview}), thereby enabling dynamic traversal of narrow paths. Our overarching objective is to achieve dynamic walking on a flexible rope or an extremely narrow path, with the findings presented here marking a significant stride toward this ambition.

The structure of this work is as follows: Initially, we provide a concise overview of the hardware configuration of Husky Carbon. Subsequently, we delineate the derivations and assumptions underlying the HROM, followed by an elucidation of the optimization-based controller proposed and implemented in this study. This controller leverages approximations of the HROM within a collocation framework to compute joint and thruster commands. Lastly, we present the simulation results and conclude the work with closing remarks and prospects for future research endeavors.

\section{Quick Overview of Husky Carbon}
\label{sec:husky-overview}

Husky Carbon, is a quadrupedal robot with multi-modal capabilities, custom designed and fabricated at Northeastern University's \textit{SiliconSynapse Labs}. Husky Carbon \cite{ramezani_generative_2021}, along with Husky Beta \cite{wang_legged_2023}, Harpy \cite{liang_rough-terrain_2021,sihite_unilateral_2021,sihite_optimization-free_2021} and M4 \cite{sihite_multi-modal_2023,mandralis_minimum_2023, sihite_demonstrating_2023, sihite_dynamic_2023} join the suite of robots with multimodal capabilities. Husky Carbon, hereby referred to just as Husky, stands at 1.5 ft wide and 3 ft tall. The robot was designed and fabricated extensively using additively manufactured components. 

The hip-sagittal (HS) joint works in tandem with the knee (K) joint to maneuver the leg in the hip sagittal plane. In the interest of this research, the hip frontal (HF) joints play a very important role by being able to control the position of the foot in the frontal plane of the robot. With three motors to control the above three joints, all the 12 joints on the robot are actuated by T-motor Antigravity 4006 brushless motors, with the motor output transmitted through a Harmonic drive. The Harmonic drives are chosen for their precise transmission, low backlash, and back-drivability. The motor and gearbox housings along were embedded in the housing during the printing process making the robot's legs significantly lightweight.

To power the 12 actuators, the robot uses 12 ELMO gold twitter solo amplifiers and a Speedgoat realtime machine that sends control commands and also obtains position feedback from the drives using incremental magnetic encoders. The amplifiers are isolated away from the actuators and are separated into 2 racks of 6 amplifiers each and then mounted onto either side of the robot. The host PC running MATLAB is connected to the Realtime machine using EtherCAT.

The propulsion system is fitted with 4 Schubeler Electric Ducted fans (EDF) that provide approximately 8 kgf of thrust in total. It is built with a lightweight hexagonal aluminum composite structure sandwiched between carbon fiber plates, the design of which is extensively delineated in \cite{pitroda_dynamic_2023}.

\section{Husky Reduced-Order Model (HROM)}
\label{sec:hrom}

\begin{figure*}
    \centering
    \includegraphics[width = 0.8\linewidth]{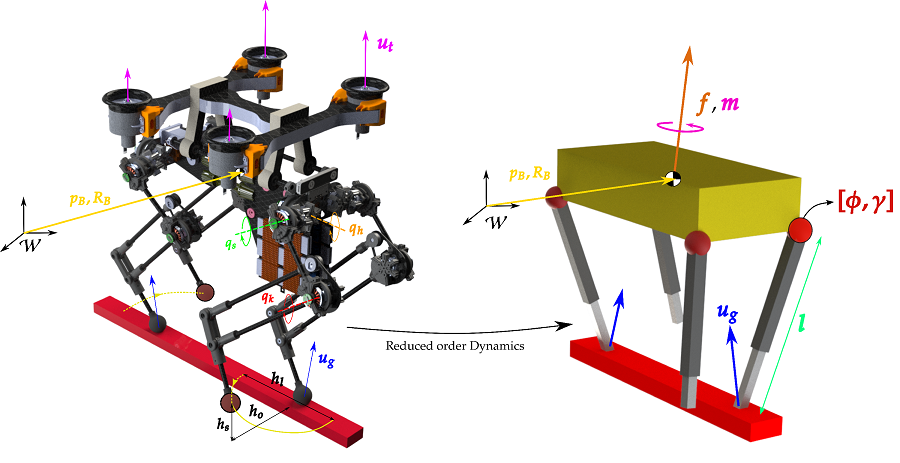}
    \caption{ Illustrates the Husky full-fidelity model vs. the reduced order model with the model parameters used in the derivations in Section~\ref{sec:hrom}.
    }
    \label{fig:hrom-image}
\end{figure*}

The equations of motion of the HROM can be derived using the energy based Euler-Lagrange dynamics formulation. As shown in Fig.~\ref{fig:hrom-image}, the positions of the leg ends are defined as functions of the spherical joint primitives, namely $\phi$ and $\gamma$, along with the length of the leg $l$. The pose of the body can be defined using $\bm{p}_B \in \mathbb{R}^3$, and Z-Y-X Euler angles $\bm{\Phi}_B$. The rotation matrix can also then be defined from the Euler matrix as $R_B$. The generalized coordinates of the robot body can then be defined as follows:
\begin{equation}
\bm{q} = [\bm{p}_B^\top ,\bm{\Phi}_B^\top ]^\top,
\label{eq: body states}
\end{equation}
and the leg states of the robot can be defined as,
\begin{equation}
\begin{aligned}
\bm{q}_L &= [\dots, \phi_i,\gamma_i,l_i, \dots]^\top, \\
i &\in \mathcal{F},
\label{eq: leg states}
\end{aligned}
\end{equation}
where $\mathcal{F} =\left\{FR, HR , FL , HL\right\}$ represents the respective legs and thrusters. The position of the foot can then be determined using the forward kinematics equations shown:
\begin{equation}
\begin{aligned}
    \bm{p}_{F i} &= \bm{p}_{B} + R_{B} \bm{l}_{h i}^{B} + R_{B} \bm{l}_{f i}^{B} \\
    \bm{l}_{f i}^{B} &= R_{y}\left(\phi_{i}\right) R_{x}\left(\gamma_{i}\right)
    \begin{bmatrix} 
    0, & 0, & -l_{i}
\end{bmatrix}^\top
\label{eq:foot_pos}
\end{aligned}
\end{equation}
The positions of the thrusters are defined as $\bm{p}_{ti}$ with respect to the body. The superscript $B$ denotes a vector defined in the body frame, while the rotation matrix $\bm{R_B}$ represents the rotation of a vector from the body frame to the inertial frame. Since the legs are considered massless, the kinetic and potential energies of the HROM can be calculated using the equations shown below:
\begin{equation}
\begin{aligned}
    \mathcal{K} &= \left( \frac{1}{2} \dot{\bm{p}}_B m_B \dot{\bm{p}}_B^\top +\bm{\omega}_B^B I_B \bm{\omega}^{B\top}_B \right) \\
    \mathcal{V} &= -m_B \bm{p}_B^\top \bm{g} \\
    \mathcal{L} &= \mathcal{K}-\mathcal{V},
\label{eq:LKV}
\end{aligned}
\end{equation}

\noindent where $\bm{\omega}_B^B$ represents the body angular velocity in the body frame, and $\bm{g}$ denotes the gravitational acceleration vector. The angular velocity of the body can be found as a function of the rate of change of Euler angles using the Euler rate matrix E, 
\begin{equation}
\bm{\omega}_B^B = E \dot{\bm{\Phi}}
\end{equation}

If the generalized velocities of the system is defined as $\bm{v}=~ \left[\dot{\bm{p}}_B, \bm{\omega}_B\right]$, the Lagrangian of the system can be calculated as $\mathcal{L} = \mathcal{K} - \mathcal{V}$, and the dynamic equation of motion can be derived using the Euler-Lagrangian method as follows:
\begin{equation}
    \textstyle \frac{d}{dt}\left(\frac{\partial{\mathcal{L}}}{\partial{\bm{v}}}\right)- \frac{\partial \mathcal{L}}{\partial{\bm{q}}} = \bm{\Gamma},
\label{eq:euler-lagrangian}
\end{equation}
where $\bm{\Gamma}$ is the sum of all generalized torques and forces respectively. The dynamic system accelerations can then be solved to obtain the into the following standard form:
\begin{equation}
\begin{gathered}
    \bm D(\bm{q})\dot{\bm{v}} + \bm C(\bm{q}, \bm{v})\bm{v} + \bm G(\bm{q}) = \Sigma_{i \in \mathcal{F}} \left[ \bm B_{gi}\bm{u}_{gi} \right] + \bm{u}_t \\
    \bm B_{gi} = \frac{\partial{\dot{\bm{p}}_{f,i}}}{\partial{\bm{v}}},
\end{gathered}
\label{eq:manipulator eq}
\end{equation}

\noindent where $\bm D$ is the mass-inertia matrix, $\bm C$ contains the Coriolis vectors and gravitational vectors are defined in $\bm G$, and $\bm B_{gi} \bm u_{gi}$ represent the generalized force due to the GRF (Ground Reaction Forces) $\bm u_{gi}$ acting on the foot $i$. The term $\bm{u}_t \in \mathbb{R}^6$ represents the external wrench acting on the COM of rigid body of the HROM i.e $\bm u_t = [\bm{f}^{\top},\bm{m}^{\top}]^{\top}$ and $\bm{f},\bm{m}$ are the forces and moments that form the wrench. 

The vector $\bm u_t$ represents the actions exerted by the four thrusters, which is fprmed by condensing the thruster forces into a wrench. The HROM's legs are then driven by setting the joint variable's accelerations to track desired joint states. The joint inputs are defined as follows
\begin{equation}
    \ddot{\bm q}_L = \bm u_L,
    \label{joint-inputs}
\end{equation}
\noindent where, $\bm{u}_L$ forms the control input to the system in the form of the leg joint state accelerations. 

The full system of equations can then be derived from equation \ref{eq:manipulator eq} and equation \ref{joint-inputs} as follows:
where $\bm q_d = [\bm q^\top, \bm q_L^\top]^\top, \bm{v}_d = \left[\bm{v}^\top, \bm{\dot{q}}_L^\top\right]^\top$, and $\bm{x}$ is obtained by combining both the dynamic and massless leg states and their derivatives to form the full system states. Finally, $\bm{u}$ is a vector of the $m$ inputs, which include the thrust wrench and the leg inputs. 

The GRF is modeled using a compliant ground model and Stribeck friction model, defined as follows:
\begin{equation}
\Sigma_{GRF}:\left\{\begin{aligned}
    \bm u_{gi} & = \begin{cases} \, 0 &  \mbox{if } z_i > 0  \\
    \, [u_{gi,x},\, u_{gi,y},\, u_{gi,z}]^\top & \mbox{else} \end{cases} \\
    u_{gi,z} &= -k_{gz} z_i - k_{dz} \dot{z}_i \\
    u_{gi,x} &= - s_{i,x} u_{gi,z} \, \mathrm{sgn}(\dot{x}_i) - \mu_v \dot{x}_i \\
    s_{i,x} &= \left(\mu_c - (\mu_c - \mu_s) \mathrm{exp} \left(-|\dot{x}_i|^2/v_s^2  \right) \right),
\end{aligned}\right.
\label{eq:ground-model}
\end{equation}
where $x_i$ and $z_i$ represent the $x$ and $z$ positions of foot $i$, respectively. $k_{gz}$ and $k_{dz}$ are the spring and damping coefficients of the compliant surface model, respectively. $u_{gi,x}$ and $u_{gi,y}$ denote the ground friction forces in the respective directions. $\mu_c$, $\mu_s$, and $\mu_v$ stand for the Coulomb, static, and viscous friction coefficients, respectively, and $v_s > 0$ represents the Stribeck velocity. The derivations of $u_{g,y}$ follow similarly to those of $u_{g,x}$.

\section{Controls}
\label{sec:ctrl}

To solve this controls problem, i.e., find $\bm u_t$ from Eq.~\ref{eq:manipulator eq}, we consider the following cost function given by

\begin{equation}
    \centering
    \begin{aligned}
    J =& \sum_{k=1}^{N} \bm{x}_{e,k}^{\top}Q\bm{x}_{e,k} + \bm{u}_{t,k}^{\top}R\bm{u}^{}_{t,k},
    \end{aligned}
\end{equation}

\begin{figure*}[t]
    \centering
    \includegraphics[width = 0.8\linewidth]{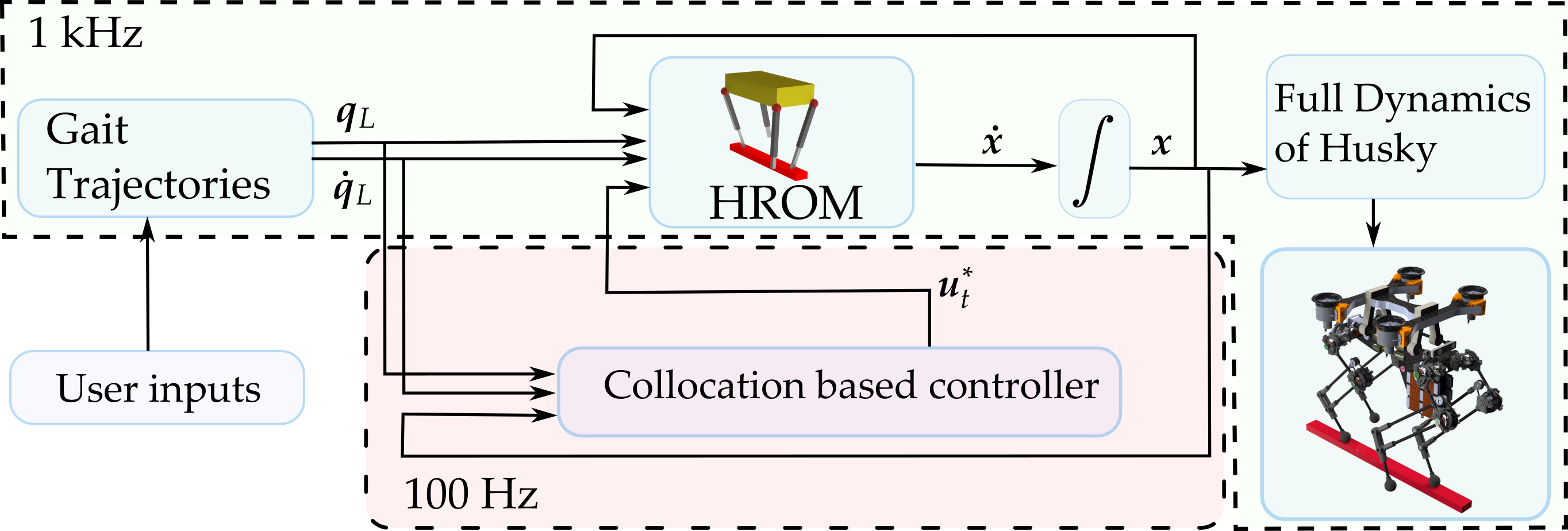}
    \caption{The proposed control system flowchart. The user inputs include forward velocity reference and gait parameters.}
    \label{fig:ctrl-diagram}
\end{figure*}

\noindent where $\bm{x}_e \in \mathbb{R}^3$ is an error term that calculates the error of the body pose in the form of Euler angles calculated from the body transformation matrix ${R_B}$. $Q,R \in \mathbb{R}^{3}$ are positive definite matrices containing weights of the corresponding error terms.
The cost function $J$ is governed by the model given by Eq.~\ref{eq:manipulator eq}. We perform temporal (i.e., $t_i,~i=1, \ldots, n, \quad 0 \leq t_i \leq t_f$) discretization to obtain the following system of equations,
\begin{equation}
\dot{\bm{x}}_i=\bm{f}_i(\bm{x}_i,\bm{u}_i), \quad i=1, \ldots, n, \quad 0 \leq t_i \leq t_f,
\label{eq:discrete-hrom-model} 
\end{equation}

\noindent where $\bm{x}_i$ embodies the values of the state vector $\bm x$ at i-th discrete time. And, $\bm{u}_i$ the thruster commands in the form of the wrench at the i-th sample time. $\bm{f}_i$ denotes the governing dynamics of HROM at i-th discrete time.

We stack all of the discrete values from $\bm{x}_i$ and $\bm{u}_i$ in the vectors $\bm{X} = \left[\bm{x}^\top_1(t_1), \ldots, \bm{x}_n^\top(t_n)\right]^\top$ and $\bm U = \left[\bm{u}^\top_1(t_1), \ldots, \bm{u}^\top_n(t_n)\right]^\top$.

We consider 2$n$ boundary conditions at the boundaries of $n$ discrete periods to ensure continuity, given by
\begin{equation}
    r_i\left(\bm{x}(0), \bm{x}\left(t_f\right), t_f\right)=0, \quad i=1, \ldots, 2 n
\end{equation}

Since we have $m$ entries in $\bm{u}$, we consider $m$ inequality constraints to ensure the thruster forces remain inside the constrained admissible set. $\bm{g}_i$ is give
\begin{equation}
    \bm{g}_i(\bm{x}(t_i), \bm{u}(t_i), t_i) \geq 0, \quad i=1,\dots,m \quad 0 \leq t_i \leq t_f
\end{equation}
To approximate the nonlinear dynamics from HROM, we employ a method based on polynomial interpolations. This method extremely simplifies the computation efforts. Consider the $n$ time intervals, as defined previously and given by 
\begin{equation}
0=t_1<t_2<\ldots<t_n=t_f
\end{equation}
 We stack the states $\bm{x}_i$ and input terms $\bm{u}_i$ given by $\bm{X} = \left[\bm{x}^\top_1,\dots,\bm{x}^\top_n\right]^\top$ and $\bm{U} =\left[\bm{u}_1^\top,\dots,\bm{u}_n^\top\right]^\top$ from the HROM at these discrete times into a single vector denoted by $\mathcal{Y}$ and form a decision parameter vector that the optimizer finds at once. Additionally, we append the final discrete time $t_f$ as the last entry of $\mathcal{Y}$ so that walking speed too is determined by the optimizer.
\begin{equation}
\mathcal{Y}=\left[\bm{x}^\top_1,\dots,\bm{x}^\top_n,\bm{u}_1^\top,\dots,\bm{u}_n^\top,t_f \right]^\top
\end{equation}
We approximate the input vector $\bm{u}_i(t_i)$ at time $t_i \leq t<t_{i+1}$ as the linear interpolation function $\bm{\tilde u}$ between $\bm{u}_i(t_i)$ and $\bm{u}_{i+1}(t_{i+1})$ given by 
\begin{equation}
\tilde{\bm{u}}= \bm{u}_i\left(t_i\right)+\frac{t-t_i}{t_{i+1}-t_i}\left( \bm{u}_{i+1}\left(t_{i+1}\right)-\bm{u}_i\left(t_i\right)\right)
\end{equation}

We interpolate the states $\bm{x}_i(t_i)$ and $\bm{x}_{i+1}(t_{i+1})$ as well. However, we use a nonlinear cubic interpolation, which is continuously differentiable with $\dot{\tilde{\bm{x}}}(s)= \bm{f}(\bm{x}(s), \bm{u}(s), s)$ at $s=t_i$ and $s=t_{i+1}$. 

To obtain $\tilde{\bm{x}}$, we formulate the following system of equations:
\begin{equation}
    \begin{aligned}
\tilde{\bm{x}}(t) &=\sum_{k=0}^3 c_k^j\left(\frac{t-t_j}{h_j}\right)^k, \quad t_j \leq t<t_{j+1}, \\
c_0^j &=\bm{x}\left(t_j\right), \\
c_1^j &=h_j \bm{f}_j, \\
c_2^j &=-3 \bm{x}\left(t_j\right)-2 h_j \bm{f}_j+3 \bm{x}\left(t_{j+1}\right)-h_j \bm{f}_{j+1}, \\
c_3^j &=2 \bm{x}\left(t_j\right)+h_j \bm{f}_j \bm{x}\left(t_{j+1}\right)+h_j \bm{f}_{j+1}, \\
\text { where } \bm{f}_j &:=\bm{f}\left(\bm{x}\left(t_j\right), \bm{u}\left(t_j\right)\right), \quad h_j:=t_{j+1}-t_j .
\end{aligned}
\label{eq:cubic-lobatto}
\end{equation}
The interpolation function $\tilde{\bm{x}}$ utilized for $\bm{x}$ needs to fulfill the continuity at discrete points and at the midpoint of sample times. 

By examining Eq.~\ref{eq:cubic-lobatto}, it is evident that the derivative terms at the boundaries $t_{i}$ and $t_{i+1}$ are satisfied. Hence, the only remaining constraints in the nonlinear programming problem are the collocation constraints at the midpoint of $t_i-t_{i+1}$ time intervals, the inequality constraints at $t_i$, and the constraints at $t_1$ and $t_f$, all of which are included in the optimization process. We address this optimization problem using MATLAB's fmincon function. The overall proposed control system flowchart is depicted in Fig. ~\ref{fig:ctrl-diagram} along with the user-inputs into the system, which involves the forward velocity reference and gait parameters. 

\begin{figure}
    \centering
    \includegraphics[width = \linewidth]{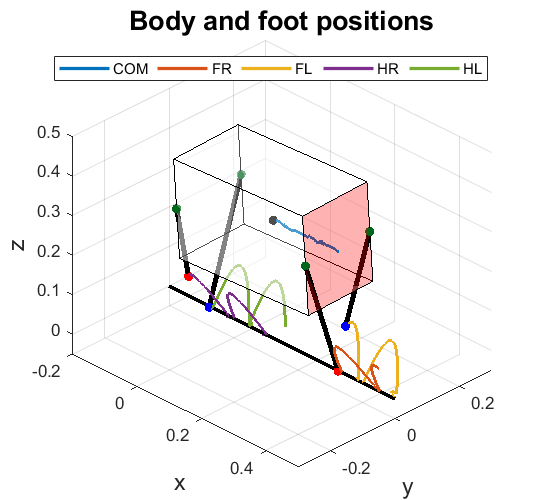}
    \caption{Illustrates the stick-diagram of HROM traversing a narrow path with 3D CoM and leg-end trajectories}
    \label{fig:3D-traj}
\end{figure}

  \section{Results}
  \label{sec:results}

This simulation was performed in the MATLAB environment using a computer with an Intel core i7 processor and utilized the HROM framework, supported by MATLAB animations, to model and analyze the system's behavior. A fourth-order Runge Kutta integrator was used to march the ODE forward. Basic heuristics were then applied to determine gaits for a straight path, considering a specified forward velocity and step time, with a simulation duration of 3.5 seconds. Bezier control points were then generated based on user defined inputs to create desired trajectories for both swing and stance phases, employing seven control points. The simulation adopted a bipedal locomotion pattern with two-point contact, where diagonally opposite leg pairs synchronized while the remaining pair operated out of phase.

\begin{figure}[h]
    \centering
    \includegraphics[width = \linewidth]{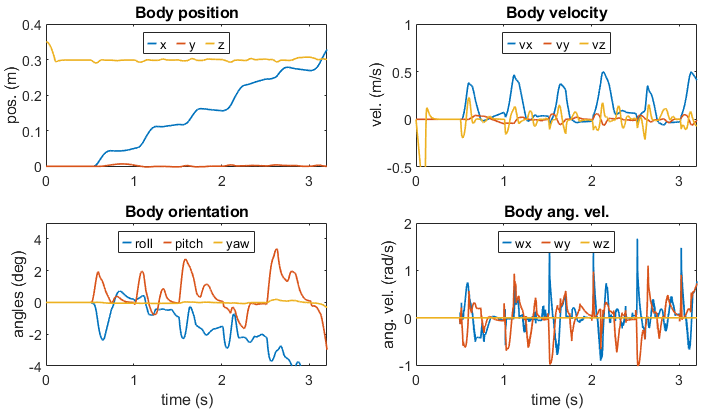}
    \caption{Illustrates body position and orientation $\bm q$ and $\bm{\dot{q}}$ under $\bm u$ and governing dynamics given by Eq.~\ref{eq:manipulator eq}}
    \label{fig:body-states}
\end{figure}

\begin{figure}[h]
    \centering
    \includegraphics[width = \linewidth]{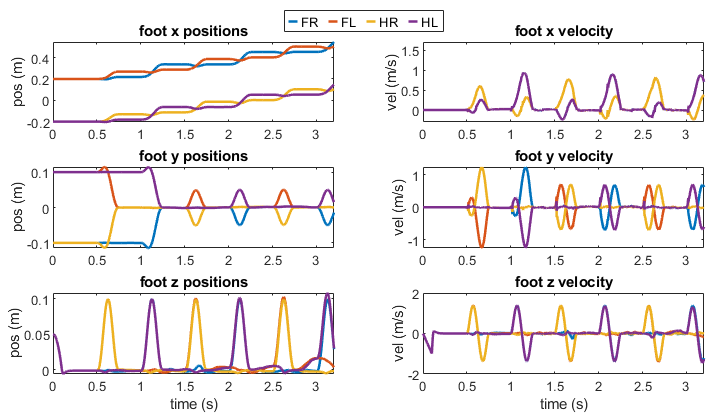}
    \caption{Illustrates leg end positions and velocities}
    \label{fig: foot-states}
\end{figure}

Figure~\ref{fig: foot-states} illustrates foot positions in the global reference frame during locomotion. Initially, all legs moved from neutral positions toward the body's center for narrow-path walking. Subsequently, a leg pair swung outward to avoid stance legs before returning inward for the stance phase followed by a small pause period. Figure~\ref{fig: foot-states} plots, particularly foot y positions, confirmed these movements.

\begin{figure}
    \centering
    \includegraphics[width = \linewidth]{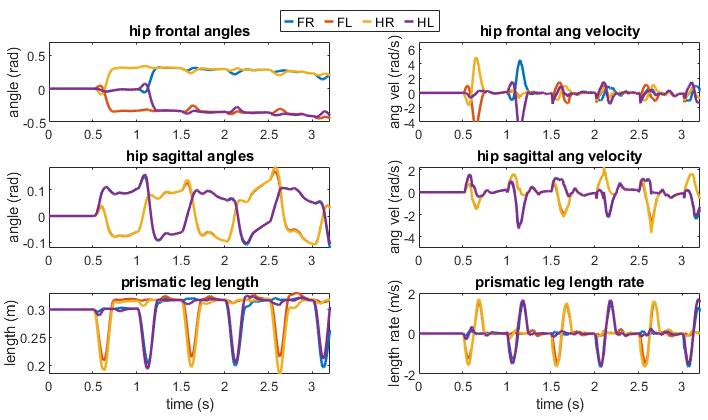}
    \caption{Illustrates leg joint angles and agular speeds}
    \label{fig: joint-traj}
\end{figure}

Consequently, Fig.~\ref{fig: joint-traj} depicts the joint positions and their respective angular velocities. The plots show the hip frontal angle moving during the first step to facilitate foot positioning in the frontal plane. In Fig.~\ref{fig:3D-traj}, we can see the resulting 3D trajectory of the foot-ends, exhibiting the narrow path gait of the foot end and the body moving forward as a result. 

\begin{figure}[t]
    \centering
    \includegraphics[width = 0.8\linewidth]{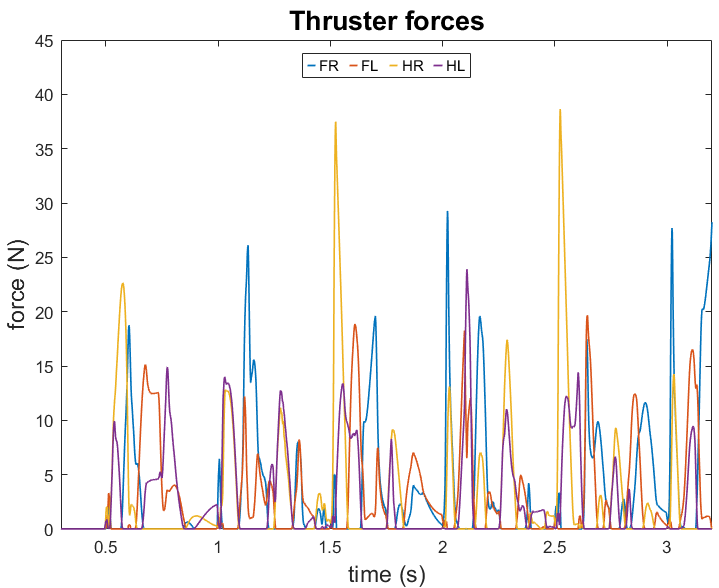}
    \caption{Shows thruster commands obtained by the controller following stable narrow-path dynamic walking.}
    \label{fig: thruster-forces}
\end{figure}

In Fig.~\ref{fig:body-states}, it can be observed that the body COM position moves forward during each gait cycle, with stable y and z positions. During the period of the simulation the body moves forward about 0.3 meters in 3.5 secs, equating to about an average velocity of 0.1 m/s. To control the attitude of the body, a thrust wrench located at the COM is used as explained in Section \ref{sec:hrom}. The decomposed wrench acting as four separate normal forces in the body floating body frame in shown in Fig.~\ref{fig: thruster-forces}. For the purposes of simulation and to validate the theory that the body attitude can indeed be controlled by a wrench, the thrust forces were generated by the controller using the pose error from Euler angles obtained from the rotation matrix.

Figure~\ref{fig:body-states} also revealed that for the duration of the simulation, the body's attitude stays stable, with small angular velocities. The thrust forces generated by the controller also reveals that for the majority of the duration, the  generated thruster forces are within the threshold of the acceptable limits of the EDFs present on the physical robot. This can be further improved with the implementation of the optimal controller described in \ref{sec:ctrl}.

\begin{figure}
    \centering
    \includegraphics[width = \linewidth]{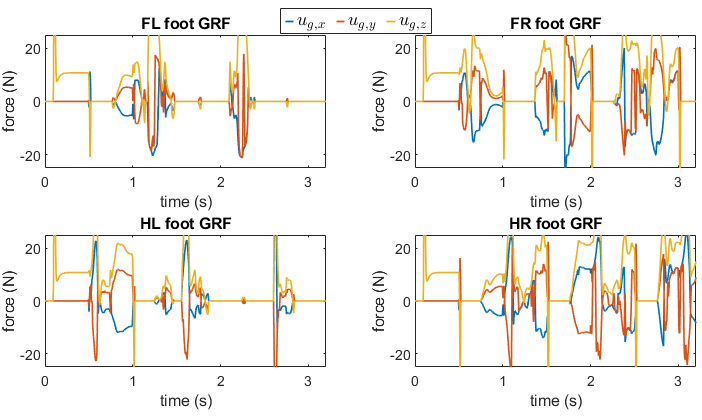}
    \caption{Illustrates ground contact forces $\bm u_{gi}$ obtained from Eq.~\ref{eq:ground-model}}
    \label{fig: GRF-plot}
\end{figure}

The GRF for each leg generated from a compliant ground model calculated as a function of the vertical foot position from Eq.~\ref{eq:ground-model}, and can be seen in Fig.~\ref{fig: GRF-plot}. The stiffness of the ground was taken to be 8000 N/m and the damping was taken to be 250 Ns/m. The friction coefficients for the Coloumb, Stribeck and viscous friction were taken as 0.5, 0.6,and 0.8 respectively. The Stribeck velocity was chosen to be 0.01 m/s. With the given parameters,  the simulation demonstrated minimal slippage of the leg relative to the ground, indicating stability during legged locomotion.

\section{Concluding Remarks}
\label{sec:conclusion}

In this paper, we propose a control framework for utilizing a reduced-order model to achieve narrow path walking with thruster stabilization of Husky Carbon. Husky Carbon, a quadrupedal robot with multi-modal capabilities, was custom designed and fabricated at Northeastern University's \textit{SiliconSynapse Labs}. Alongside Husky Beta, Harpy, and M4, Husky Carbon enriches the suite of robots with multimodal capabilities. Husky Carbon, referred to simply as Husky, measures 1.5 ft wide and 3 ft tall and was extensively designed and fabricated using additively manufactured components.

We utilize the reduced-order model in conjunction with an optimization-based controller employing polynomial approximation to determine the state and input values for controlling Husky's dynamic walking gaits on a narrow path. Our assumption of a compliant ground model links our narrow path locomotion to other scenarios such as slacklining, suggesting that our proposed control design paradigm can be extended to other narrow path locomotion scenarios.

Future work will involve enhancing the accuracy of our simulators and conducting real-world testing of the proposed collocation-based optimal controller. Furthermore, we aim to validate our results on higher fidelity simulations of Husky, where constraints for gaits must also be considered to account for potential self-collisions in walking scenarios, as the legs move close to the body. 

\printbibliography
\end{document}